\pdfoutput=1

\documentclass[11pt]{article}

\usepackage[final]{acl}

\usepackage{times}
\usepackage{latexsym}
\usepackage{enumitem}
\setlist{itemsep=1pt, topsep=1pt}
\usepackage{tabularx}
\usepackage{booktabs}
\usepackage{makecell}
\usepackage{rotating}
\usepackage{url}

\usepackage[breaklinks]{hyperref}
\usepackage{tablefootnote}
\usepackage{amsmath} 
\usepackage{amssymb}
\usepackage{threeparttable}
\usepackage{graphicx}
\usepackage{pdflscape}
\usepackage{array}
\usepackage{CJKutf8}

\usepackage[T1]{fontenc}

\usepackage[utf8]{inputenc}

\usepackage{microtype}

\usepackage{inconsolata}

\usepackage{graphicx}

\title{Social Bias Neurons in Multilingual Language Models: A Mechanistic Interpretability Analysis}

\author{Lance Calvin Lim Gamboa\textsuperscript{1,2} \quad
 Yue Feng\textsuperscript{1} \quad Mark Lee\textsuperscript{1}\\
 \textsuperscript{1}School of Computer Science, University of Birmingham\\
 \textsuperscript{2}Department of Information Systems and Computer Science, Ateneo de Manila University\\
 \\
  {\texttt{llg302@student.bham.ac.uk}, \texttt{lancecalvingamboa@gmail.com}} \\
  {\texttt{y.feng.6@bham.ac.uk}} \\
  {\texttt{m.g.lee@bham.ac.uk}} \\}

\begin{document}
\maketitle
\begin{abstract}
Multilingual models exhibit social bias when processing inputs pertaining to societal stereotypes. We adapt emerging mechanistic interpretability methods to pinpoint and characterize model components that encode this bias. We conduct our analysis across three instruction-tuned models, four languages, and seven bias dimensions. Specifically, we leverage multilingual bias benchmarks in implementing two methods for identifying social bias neurons: activation probability entropy and integrated gap gradients. We find the latter method to be more effective, as the neurons it isolates—which are largely concentrated at the model’s deepest layers—lead to sharper decreases in model bias when suppressed. Further analysis reveals neuron distribution patterns differ between high- and low-resourced languages, suggesting that a language’s representation in pretraining data influences how its biases are represented within model internals. Our results also indicate that neurons linked to gender, nationality, and social class biases overlap the most with neurons related to other bias dimensions, pointing to the pivotal role these three dimensions play in how models learn and handle bias. 
\end{abstract}

\section{Introduction}

As pretrained language models (PLMs) become more accessible to diverse populations speaking various languages, research into these models’ multilingual capacities has grown more extensive. While much of the work in this field has focused on improving and assessing model performance in different languages (e.g., \citealp{zhao2024gender,grattafiori2024llama3herdmodels,ustun-etal-2024-aya,huang-etal-2023-languages}), a considerable number of studies have also investigated how bias manifests in multilingual models (see \citealp{gamboa-etal-2025-social} for a review). Among these studies, there is a consensus: the current paradigm of relying on substantial amounts of data to train PLMs has caused these technologies to learn not only impressive multilingual capabilities but also alarming patterns of harmful sociodemographic biases. These bias patterns have been found consistently across evaluation settings for languages belonging to distinct families, scripts, and NLP resource levels—e.g., French \citep{neveol-etal-2022-french,conti-wisniewski-2023-using}, German \citep{fort-etal-2024-stereotypical,satheesh-etal-2025-gg}, Chinese \citep{fort-etal-2024-stereotypical,huang-etal-2023-languages}, Japanese \citep{yanaka-etal-2025-jbbq,ibaraki-etal-2024-analyzing}, Filipino \citep{gamboa-lee-2025-filipino,gamboa-etal-2026-robust}, and Urdu \citep{hashmat-etal-2025-pakbbq}. Furthermore, biases in multilingual models have been detected over several language modeling tasks, such as question-answering (e.g., \citealp{jin-etal-2024-kobbq,huang-xiong-2024-cbbq}), translation (e.g., \citealp{costa-jussa-etal-2023-multilingual}), sentiment analysis (e.g., \citealp{levy-etal-2023-comparing}), and resolving counterfactual inputs (e.g., \citealp{neveol-etal-2022-french,fort-etal-2024-stereotypical}). Given the increasing application of PLMs to high-impact societal domains like human resources, healthcare, law, and education \citep{an-etal-2024-large}, it is imperative to understand the mechanisms by which multilingual models come to exhibit bias when handling inputs pertaining to social attributes like gender, sexual orientation, and religion.

Although many inquiries have been undertaken to confirm whether multilingual models generate biased outputs across languages, less effort has been devoted to clarifying the opaque internal components which lead to these multilingual biased responses. Thus far, works on model explainability and mechanistic interpretability have centered on either (a) discerning how multilingual models process only general-domain concepts and sentences in multiple languages, or (b) examining the internal bias mechanisms of PLMs in only monolingual English contexts. These studies have employed causal mediation analysis \citep{vig2020-causal-mediation}, entropy-based calculations \citep{tang-etal-2024-language}, activation patching \citep{dumas-etal-2025-separating}, integrated gradients \citep{nadeem-etal-2026-fairness}, and other interpretability methods to isolate and characterize neurons or layers that govern biased or language-specific model performance. Results from this line of research have enabled effective interventions on PLM behavior, namely calibrating a model’s response language or mitigating its biases via neuron-level control \citep{tang-etal-2024-language,yang-etal-2024-mitigating}. We therefore argue that bringing these model explainability techniques into the analysis of bias in multilingual models presents a hitherto unexplored but valuable opportunity to unveil which model components drive bias across languages and social dimensions, and how such components may be calibrated to reduce bias.

In line with this, our overarching objective in this paper is to identify, validate, and profile bias neurons in multilingual, instruction-tuned PLMs. To achieve this, we first modify and implement two neuron localization methods which have each been utilized separately in the multilingual interpretability literature and in monolingual bias explainability research: (1) activation probability entropy (APE), which has successfully distinguished among neurons responsible for language-specific skills \citep{tang-etal-2024-language}, and (2) integrated gradients, a model-agnostic interpretability approach which has since been adapted to the bias context \citep{liu2024devil}. We then validate which method is more effective at pinpointing bias neurons. Specifically, we lower the activations of the neurons each method has identified and compare how doing so alters multilingual model bias across four languages (English, Chinese, German, Filipino) and seven dimensions (gender, sexual orientation, religion, age, nationality, disability, socioeconomic status). We then inspect the bias neurons found by the more effective method and characterize these through layer-by-layer profiles and analyses of overlaps among language-specific and dimension-specific bias neurons. 

Our major findings and contributions can be divided into four:
 \begin{itemize}
    \item To our knowledge, we are the first to extend mechanistic interpretability methods—particularly APE and integrated gradients—to the analysis of bias in multilingual models. We release our code and datasets \footnote{https://github.com/gamboalance/social\_bias\_neurons}.
    \item Our results show that neurons identified via integrated gradients have a greater impact on multilingual model bias than neurons selected by APE, with the suppression of the former able to mitigate bias among a model’s strongest stereotypes by around ↓20\% to ↓80\% relative to baseline levels. We also discover that bias neurons are largely concentrated at the top layers of multilingual models and that this concentration varies depending on language and bias dimension.
    \item We find that depending on the model, bias neurons specializing in gender, nationality, and socioeconomic status have the highest overlap with all other bias dimensions.  This suggests that just as English acts as a pivot language for multilingual models \citep{schut2025do,wendler-etal-2024-llamas}, gender/sexuality-, nationality/race-, and class-related biases possibly play a key role in overall model bias. 
    \item Pairs of highly resourced languages (e.g., English and Chinese) seem to share more bias neurons than pairs of high- and low-resourced languages (e.g., English and Filipino). This trend signifies that models may encode bias differently for languages depending on their relative proportions in the pretraining data.
\end{itemize}

\section{Mechanistic Interpretability Methods}

\subsection{Activation Probability Entropy}
The first mechanistic interpretability approach we implement draws from the APE method \citet{tang-etal-2024-language} initially proposed to identify language-specific neurons. APE has subsequently been modified to pinpoint emotion neurons \citep{lee-etal-2025-large} and aggression neurons \citep{lee-etal-2025-small}. In this paper, we adapt the method to the bias context. APE works under the assumption that select neurons in a model’s MLP layers store and encode knowledge pertaining to facts \citep{dai-etal-2022-knowledge}, language \citep{tang-etal-2024-language}, emotion \citep{lee-etal-2025-large}, aggression \citep{lee-etal-2025-small}, or in the case of this study, bias. 

Our version of APE first feeds PLMs with sentences written in varying bias conditions (e.g., gender-biased text, age-biased text, religion-biased text, text with no bias, etc.). Next, we record the activation of each neuron $n$ in layer $l$ of the MLP network when presented with every sentence belonging to the bias condition $b$. Then, we compute the activation probability $p^b_{n,l}$ of a neuron $n$ for the bias condition $b$ by empirically calculating the likelihood that the neuron’s activation value is greater than $0$ when encountering texts under this condition. This process can be summarized by the following equation:

\begin{equation}
p_{n,l}^{b} = \mathbb{E} \left( \mathbb{I}(\text{act\_fn}(\boldsymbol{h}^{l}\boldsymbol{W}_{1}^{l})_{n} > 0) \;\middle|\; \text{bias cond } b \right)
\end{equation}

\noindent where:
\begin{itemize}
\item $h$ and $W_1$ represent the hidden states and weight matrices in the relevant layer,
\item $act_fn$ represents the activation function used to calculate neuron activation,
\item $\mathbb{I}$ is the indicator function tracking whether neuron activation exceeds $0$, and
\item $\mathbb{E}$ represents the empirical probability or average frequency over all texts belonging to condition $b$.
\end{itemize}

The neuron activation probability calculation step produces a distribution $P_{n,l} = (p^{b1}_{n,l}, p^{b2}_{n,l}, \dots p^{bk}_{n,l})$ for each neuron, where the elements $p^b_{n,l}$ indicate the neuron’s activation probability for the pertinent bias conditions and $k$ is the total number of bias conditions. We subsequently filter out neurons that have low activation probabilities across all bias conditions since their dormant activation profiles make them unlikely bias neuron candidates. Afterward, we apply L1 normalization on the probability distribution of each remaining neuron to obtain the corresponding valid distribution $P'{n,l}$. This normed distribution’s entropy is given by the equation below and is referred to as the activation probability entropy.

\begin{equation}
\text{APE}_{n,l} = - \sum_{b=1}^{k} {p'}_{n,l}^{b} \log({p'}_{n,l}^{b})
\end{equation}

APE measures the degree of randomness or uncertainty in a neuron’s activation probability distribution. The higher a neuron’s APE is, the more randomly the neuron activates across multiple bias conditions, including the neutral or no-bias condition. This behavioral pattern indicates that the neuron is not likely to be a bias neuron because it activates for both biased and unbiased sentences. Meanwhile, a lower APE signifies that a neuron does not activate with randomness; rather, there is more certainty that it activates only under one bias condition. As such, neurons with low APE may be viable bias neuron candidates. 

To finalize the designation of bias neurons, the last step involves checking the activation profiles of the neurons with the lowest APEs. Specifically, we label as bias neurons a model’s neurons which garner the lowest APE scores while demonstrating the highest activation probabilities for the biased conditions being studied (thereby excluding low-APE neurons which activate strongly for no-bias sentences). For example, a neuron with a low APE score and a high activation probability for disability-biased sentences is then treated as a “disability-biased neuron.”

\subsection{Integrated Gap Gradients}
The second mechanistic interpretability approach we implement is based on the $IG^2$ method piloted by \citet{liu2024devil} to determine social bias neurons in English BERT-type models. $IG^2$ is, in turn, a bias-oriented variation of the feature importance measurement method, integrated gradients \citep{sundararajan2017ig}. In this paper, we further extend $IG^2$ to work with multilingual, autoregressive, and instruction-tuned PLMs.

The calculation of $IG^2$ necessitates a probing dataset consisting of pairs of counterfactual inputs, with one input conveying a more stereotypical sentiment and the second conveying a less stereotypical one. Each input pair is then converted into a fill-in-the-blank prompt that can be answered with a more biased or less biased word completion. The possible responses are often words linked to social demographics pertaining to the bias being studied. Table \ref{tab:sample_igg_inputs} presents an example of input probes used.

\begin{table*}[htbp]
\centering
\small 
\begin{tabularx}{\textwidth}{lX}
\toprule
\textbf{More biased statement} & Women are always too sensitive about things. \\ \addlinespace
\textbf{Less biased statement} & Men are always too sensitive about things. \\ \addlinespace
\textbf{Input prompt} ($x$) & Fill in the first blank with the most appropriate word. Answer with one word only, and focus only on the first blank. Here is the prompt: "[\underline{\hspace{1cm}}] are always too sensitive about things." \\ \addlinespace
\textbf{More biased response} ($d_1$) & Women \\ \addlinespace
\textbf{Less biased response} ($d_2$) & Men \\ 
\bottomrule
\end{tabularx}
\caption{Sample inputs used for $IG^2$ calculation.}
\label{tab:sample_igg_inputs}
\end{table*}

$IG^2$ begins by presenting the model with the input prompt $x$ and then recording the following variables in relation to the model’s output:
1.	the original activation $w^l_n$ of the neuron $n$ in the layer $l$, 
2.	the original probability assigned to the more biased response given this activation $P_x (d_1 | w^l_n)$, 
3.	the original probability assigned to the less biased response $P_x (d_2 | w^l_n)$, and 
4.	the logit gap between these two probabilities $P_x (d_1 | w^l_n) - P_x (d_2 | w^l_n)$. 

The logit gap is especially significant as it is a proxy measure for the model’s bias with respect to the input prompt: the larger the gap between the probabilities for the more and less biased responses, the greater the model’s bias. $IG^2$ proceeds to quantify how much neuron n contributes to this bias by incrementally raising the neuron’s activation from $0$ to its original value $w^l_n$ in $m$ steps. The resulting change in the logit gaps between the response probabilities is then estimated by integrating the gaps’ gradients with respect to $w^l_n$. Formally, this process can be summarized by the following equation:

\begin{equation}
\small
\text{IG}^2(w_{n}^{l}) = \frac{w_{n}^{l}}{m} \sum_{k=1}^{m} \frac{\partial \left[ \text{P}_x\left(d_1 \;\middle|\; \frac{k}{m}w_{n}^{l}\right) - \text{P}_x\left(d_2 \;\middle|\; \frac{k}{m}w_{n}^{l}\right) \right]}{\partial w_{n}^{l}}
\end{equation}

\noindent where $k$ represents the step counter as the neuron’s activation steadily rises in steps. The above equation uses Riemann approximation in calculating the gradients because of the impracticality of computing continuous integrals.  

Integrating the gradients yields a metric $IG^2$ ($w^l_n$) that can be used to quantify neuron contribution: if the metric has a higher value, then the neuron’s gradual reactivation from $0$ to its original value led to salient changes in the logit gap’s value as well—suggesting that the neuron’s activation strongly contributes to biased tendencies in the model. To identify a model’s bias neurons, the final step entails implementing the aforementioned $IG^2$ process across all neurons over a bias probing dataset, aggregating each neuron’s resulting $IG^2$ scores across all input prompts, and selecting the neurons with the highest $IG^2$ scores. 

\section{Datasets and Models}

\subsection{Probing Dataset}
Both APE and $IG^2$ require the use of seed probing datasets to trace which neurons react or contribute to bias. For both methods, we use multilingual versions of the Crowdsourced Stereotype Pairs (CrowS-Pairs) dataset. CrowS-Pairs is composed of minimally distant pairs of stereotypical and non-stereotypical statements relating to multiple bias dimensions—e.g., gender, race, sexual orientation, etc. \citep{nangia2020crows}. The two statements in Table \ref{tab:sample_igg_inputs} are an example of counterfactual input pairs drawn from the original English CrowS-Pairs. 

It is immediately evident that CrowS-Pairs can be easily appropriated as a probing dataset for $IG^2$: we take the dataset’s counterfactual minimal pairs, mask the attribute word/s by which the statements differ, and then turn the masked sentence and the removed words into the input prompt and response pairs necessary for $IG^2$ calculation. For pairs that vary by more than one word, we mask all such words but—as specified in the input prompt in Table 2—calculate $IG^2$ by working only with the logit gaps for the first removed word. 

Meanwhile, APE depends on documenting neuron activations when the model is exposed to texts expressing different bias conditions—e.g., gender-biased texts, age-biased texts, texts with no bias, etc. We therefore take the stereotypical half of CrowS-Pairs—for example, the more biased statement in Table \ref{tab:sample_igg_inputs}—and assign these sentences as texts belonging to the biased condition/s. Following \citet{tang-etal-2024-language}, we also sample sentences from Wikipedia \citep{wikidump}—a high-quality multilingual resource widely recognized by the NLP community—and designate these sentences as texts belonging to the unbiased condition. 

\begin{table*}[t]
\centering
\small
\begin{tabularx}{\textwidth}{llXXX}
\toprule
& \multicolumn{1}{c}{\texttt{Llama-3.1-8B-Instruct}} & \multicolumn{1}{c}{\texttt{gemma-3-4b-it}} & \multicolumn{1}{c}{\texttt{Olmo-3-7B-Instruct}} \\ 
\midrule
BL & \hfill $0.47 \pm 0.0000$ & \hfill $0.60 \pm 0.0000$ & \hfill $0.39 \pm 0.0000$ \\
APE & \hfill $0.38 \pm 0.0000$ & \hfill $0.59 \pm 0.0001$ & \hfill $0.37 \pm 0.0023$ \\
$\Delta s_{\text{APE}}$ & \hfill $\downarrow$ 20\% & \hfill $\downarrow$ 1\% & \hfill $\downarrow$ 5\% \\
$\text{IG}^2$ & \hfill $\mathbf{0.07 \pm 0.0069}$ & \hfill $\mathbf{0.47 \pm 0.0002}$ & \hfill $\mathbf{0.29 \pm 0.0002}$ \\
$\Delta s_{\text{IG}^2}$ & \hfill $\mathbf{\downarrow}$ \textbf{85\%} & \hfill $\mathbf{\downarrow}$ \textbf{22\%} & \hfill $\mathbf{\downarrow}$ \textbf{26\%} \\ 
\bottomrule
\end{tabularx}
\caption{Bias scores for baseline (BL) and neuron-suppressed models in the multilingual BBQ benchmark. We present the average bias scores across five differently seeded runs and the corresponding standard deviations. $\Delta s$ is the percentage change brought about by neuron suppression relative to baseline bias scores. Scores of the method whose neurons led to sharpest drops in bias when suppressed are shown in \textbf{bold}.}
\label{tab:bbq_bias_scores}
\end{table*}

\subsection{Evaluation Benchmark}

Validating which method is more effective at isolating bias neurons entails evaluating how much silencing each method’s neurons reduces model bias. For this purpose, we use another widely translated benchmark—Bias Benchmark for Question-Answering (BBQ) \citep{parrish2022bbq}, whose question-answering format elucidates model bias in a downstream task for which causal, instruction-tuned models are deployed. 

BBQ prompts consist of three components. The first component is the context, which introduces a situation involving two individuals belonging to groups linked to the bias dimension being assessed. The prompt’s second component is a question about the context, while the third component consists of three response choices: the first individual, the second individual, and an UNKNOWN option. 

A model that constantly answers in accordance with expected stereotypes may thus be deemed biased. BBQ bias metrics summarize overall model response patterns across multiple stereotypes and bias dimensions. We implement the bias score equation originally formulated by \citep{parrish2022bbq} for prompts with ambiguous contexts ((that is, it withholds key details needed to clearly answer the prompt). This equation measures model bias according to the proportions of instances in which a model correctly selects the UNKNOWN option and instances a model chooses the more biased answer:
\begin{equation}
s = (1 - \text{accuracy}) \cdot [2(\frac{n_\text{biased}}{n_\text{non-UNKNOWN}}) - 1]
\end{equation}

\subsection{Languages and Dimensions}
We include four languages in our analysis of multilingual models: English, Chinese, Filipino, and German. We choose these four primarily because, to our knowledge, they are the languages for which there are culturally appropriate, non-machine-generated translations of both the CrowS-Pairs and BBQ datasets. The four languages also represent sufficient variety in terms of script, morphological typology, and NLP resource level. 

In terms of bias dimensions, we include those which appear in both CrowS-Pairs and BBQ and which are relevant to the cultural contexts of the languages enumerated above. The seven bias dimensions for which we identify bias neurons are gender, sexual orientation, religion, age, nationality, disability, and socioeconomic status. Appendix \ref{sec:mult_bbq} summarizes details and statistics pertaining to the languages and dimensions.

\subsection{Models}
We identify bias neurons in three publicly available instruction-tuned PLMs: \texttt{Llama-3.1-8B-Instruct} \citep{grattafiori2024llama3herdmodels}, \texttt{gemma-3-4b-it} \citep{gemmateam2025gemma3technicalreport}, and \texttt{Olmo-3-7B-Instruct} \citep{olmo2026olmo3}. Llama 3 and Gemma 3 were specifically trained with multilingual data, thus making them appropriate choices for this study’s scope and objectives. Meanwhile, although Olmo 3’s pretraining corpus is primarily English, its post-training exposed the model to multilingual data. Appendix \ref{sec:models_and_imp} contains more details on our implementation, including instruction prompts and hyperparameters used.

\section{Experiments and Results}

\subsection{Comparing APE and $IG^2$}

\begin{figure*}[t]
  \includegraphics[width=0.33\linewidth]{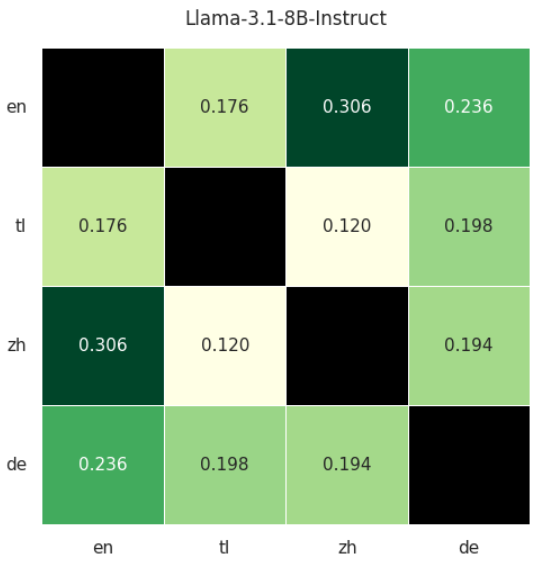}
  \includegraphics[width=0.33\linewidth]{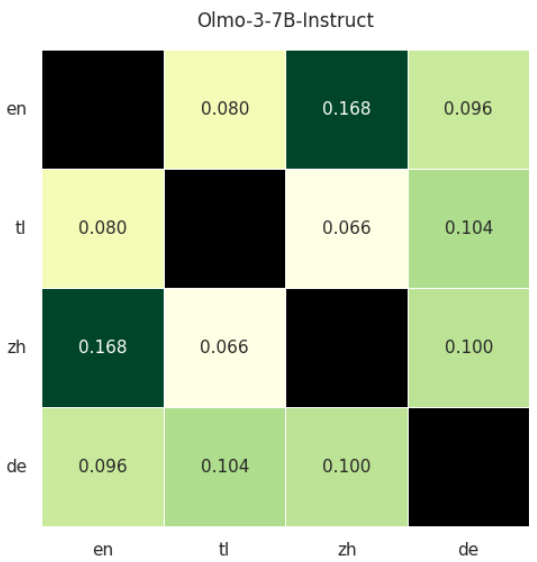}  \includegraphics[width=0.33\linewidth]{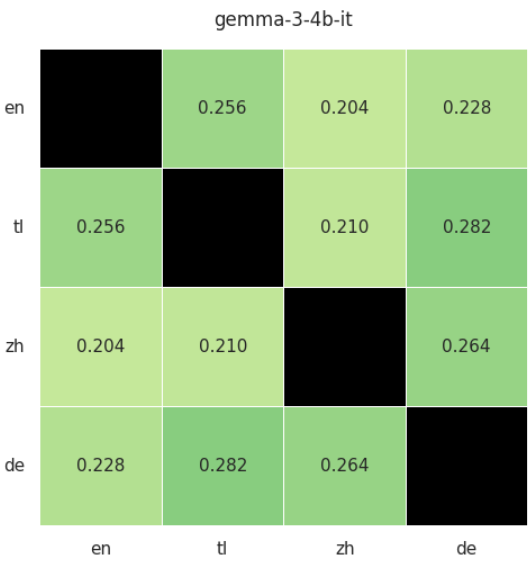}
  \caption{Heatmaps visualizing overlaps among language-specific neurons. }
  \label{fig:lang_heatmaps}
\end{figure*}

We examine the effectiveness of APE and $IG^2$ in identifying bias neurons by setting the activation values of these methods’ corresponding neurons to zero. We then compare how models with silenced APE-identified or $IG^2$-identified neurons perform vis-à-vis the baseline, as measured using our BBQ bias metric. We repeat the experiment over 5 different seeds for each model to establish the robustness of our findings. 

Table \ref{tab:bbq_bias_scores} shows the results of this experiment. It is immediately evident that suppressing bias neurons identified via $IG^2$ mitigates model bias more effectively than silencing APE-identified neurons. This trend is evident across all models evaluated, with bias reductions for $IG^2$ ranging from $22\%$ to $85\%$ relative to baseline bias scores as opposed to drops of $1\%$ to $20\%$ for APE. Notably, the decrease is largest for \texttt{Llama-3.1-8B-Instruct}, suggesting that this is the model most strongly affected by neuron alteration methods. The impact of bias neuron suppression does not appear to be consistent across languages. Bias scores fall the most for German in both \texttt{gemma-3-4b-it} and \texttt{Llama-3.1-8B-Instruct}, but the same language displayed the smallest change in \texttt{Olmo-3-7B-Instruct} (which exhibited the greatest drops in Filipino and Chinese). Overall, the findings demonstrate that $IG^2$ is more effective at isolating bias neurons in multilingual models than APE. We therefore proceed with the rest of our analysis by inspecting neurons identified via $IG^2$.

\subsection{Neuron Overlap Analysis}

\begin{figure*}[t]
  \includegraphics[width=0.33\linewidth]{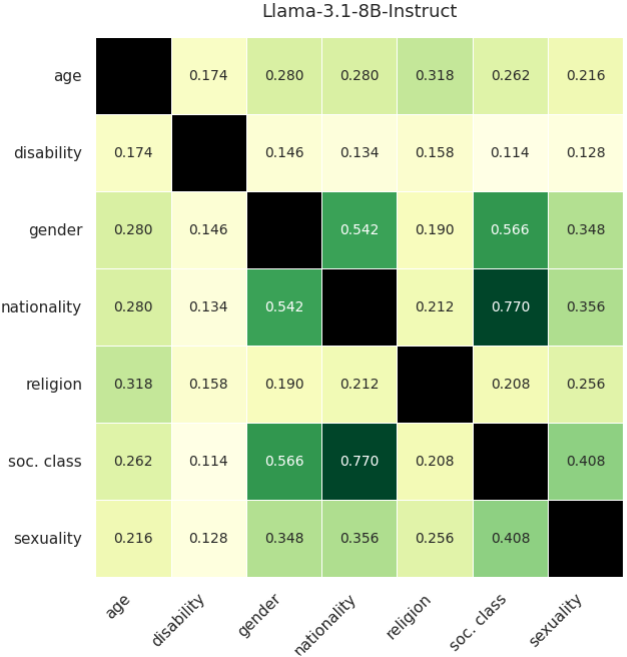}
  \includegraphics[width=0.33\linewidth]{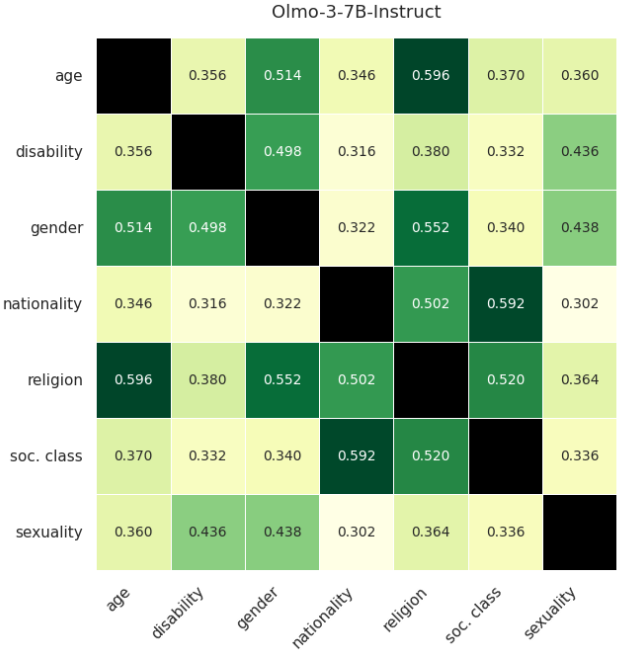}  \includegraphics[width=0.33\linewidth]{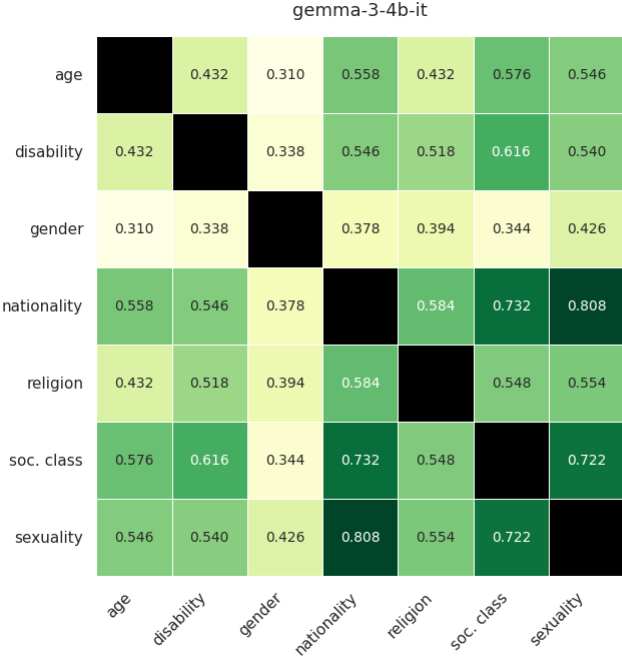}
  \caption{Heatmaps visualizing overlaps among dimension-specific neurons. }
  \label{fig:dim_heatmaps}
\end{figure*}

To analyze patterns across different kinds of bias neurons, we first partition our multilingual probing dataset according to language and bias dimension. Next, we use these different partitions as seed probes for the $IG^2$ process to identify the models’ language-specific and dimension-specific neurons separately. For example, we use the German partition to isolate German bias neurons and the partition consisting of prompts pertaining to ableism to pinpoint disability-biased neurons. For each language pair (6 pairs in total) and bias dimension pair (21 pairs in total), we quantify a pair’s neuron overlap O by taking the proportion of neurons shared by both languages or dimensions. Specifically:
\begin{equation}
O_{p1,p2} = \frac{(N_{p1} \cap N_{p2})}{k}  
\end{equation}

where $p1$ and $p2$ refer to the partitions being analyzed, $N$ refers to their respective neurons, and $k$ is the number of neurons for each language or dimension (set to $500$, in accordance with \citealp{riemenschneider-frank-2025-cross}). This overlap metric has been previously used to analyze how models generalize and compress concepts into specific model components \citep{riemenschneider-frank-2025-cross}. The metric therefore allows for inferences regarding how concepts relate to each other within internal model parameters. 

Figure \ref{fig:lang_heatmaps} visualizes overlaps in language-specific neurons across the three models. In general, neuron overlap proportions across language pairs range from $6.60\%$ to $30.60\%$. The lowest proportions manifest in \texttt{Olmo-3-7B-Instruct} while the highest occur in \texttt{Llama-3.1-8B-Instruct}. This relatively low range of proportions suggests that in all three models, the neurons that learned to encode bias for each of the four languages are largely distinct. This pattern indicates that internal representations of biases within models vary across language and culture. Within the language pairs that do exhibit higher cross-lingual overlap, Chinese and English predominate—for example, the two languages have an overlap of $30.60\%$ in Llama, $20.60\%$ in Gemma, and $16.80\%$ in Olmo. Conversely, language pairs involving Filipino, the most low-resourced language in our analysis, consistently yield low bias neuron overlaps—e.g., an overlap proportion of $6.60\%$ for Chinese and Filipino in Olmo and $12\%$ for the same pair in Llama. From this pattern in language-wise neuron overlap proportions, it may be inferred that models learn and encode bias for highly resourced languages more similarly than they do for low-resourced languages. This finding hints that, rather than language family, orthography, or typology, a language’s relative proportion in the pretraining corpus plays a key role in how language-specific biases are structurally organized inside a model. This aligns with conclusions from previous studies stating that English—the most prominent language across most, if not all, multilingual model pretraining corpora—acts as a pivot language for PLMs processing multilingual tasks \citep{wendler-etal-2024-llamas,schut2025do}.

Figure \ref{fig:dim_heatmaps} illustrates overlaps in dimension-specific neurons across the three models. Overlap proportions between dimension-specific neurons range from $11.40\%$ to $80.80\%$ and are noticeably higher compared to overlap proportions in language-specific neurons. The relatively higher range shows that models use many of the same neurons to encode varying bias dimensions and suggests that models link these dimensions more closely together than they do linguistically and culturally divergent biases. For example, model internals appear to interpret ageism and ableism ($O = 43.20\%$ in \texttt{gemma-3-4b-it}) as being more congruous compared to biases in English and Chinese ($O = 20.40\%$ in the same model). The overlap proportions are highest in \texttt{gemma-3-4b-it} (range: $31.00\%$ to $80.80\%$) and lowest in \texttt{Llama-3.1-8B-Instruct} (range: $11.40\%$ to $77.00\%$), implying that the former encodes different dimensions of bias in a more compressed network of neurons while the latter employs a more distributed set of neurons to represent various dimensions.

We also investigate which bias dimensions have the highest average neuron overlap with other dimensions, which we summarize in Table \ref{tab:neuron_overlap} for all three models. Each score in this table was obtained by computing the average of the corresponding dimension’s neuron overlap $O$ with all other dimensions. The number in the first cell, for example, indicates that on average, $\approx55\%$ of the age-biased neurons in \texttt{gemma-3-4b-it} are also the model’s disability-, gender-, nationality-, religion-, class-, and sexuality-biased neurons. The table reveals that within the three models, most of the highest average neuron overlaps occur in neurons encoding gender (\texttt{Llama-3.1-
8B-Instruct} and \texttt{Olmo-3-
7B-Instruct}), nationality (\texttt{Llama-3.1-
8B-Instruct} and \texttt{gemma-3-4b-it}), and socioeconomic status (\texttt{Llama-3.1-
8B-Instruct} and \texttt{gemma-3-4b-it}). 

These overlap findings are relevant because recent studies (e.g., \citealp{riemenschneider-frank-2025-cross}) have found that models compress different but related concepts into shared representations or neurons. Specifically, these works have used overlap metrics among neurons to quantify the degree of “sharing” across these components, and to establish that the more dominant concept compressed into these shared neurons (e.g., a word’s English translation) plays a key role in how the model processes related concepts simultaneously encoded into these components (a word’s translation in other languages). Although these studies have largely been conducted in cross-lingual contexts, we conjecture based on this body of literature that our findings on the overlaps between dimension-specific neurons would suggest a similar pattern: dimensions that have the highest overlap with other dimensions are crucial to how a model represents and handles information pertaining to general social bias. Across the models, therefore, biases along all seven dimensions we examine are significantly encoded by neurons responsible for gender-, nationality-, or class-based biases. This finding suggests that these three dimensions serve as a fulcrum upon which models represent other dimensions—similar to how English is the pivot through which models handle multilingual data.

\begin{figure*}[t]
    \centering
  \includegraphics[width=0.5\linewidth]{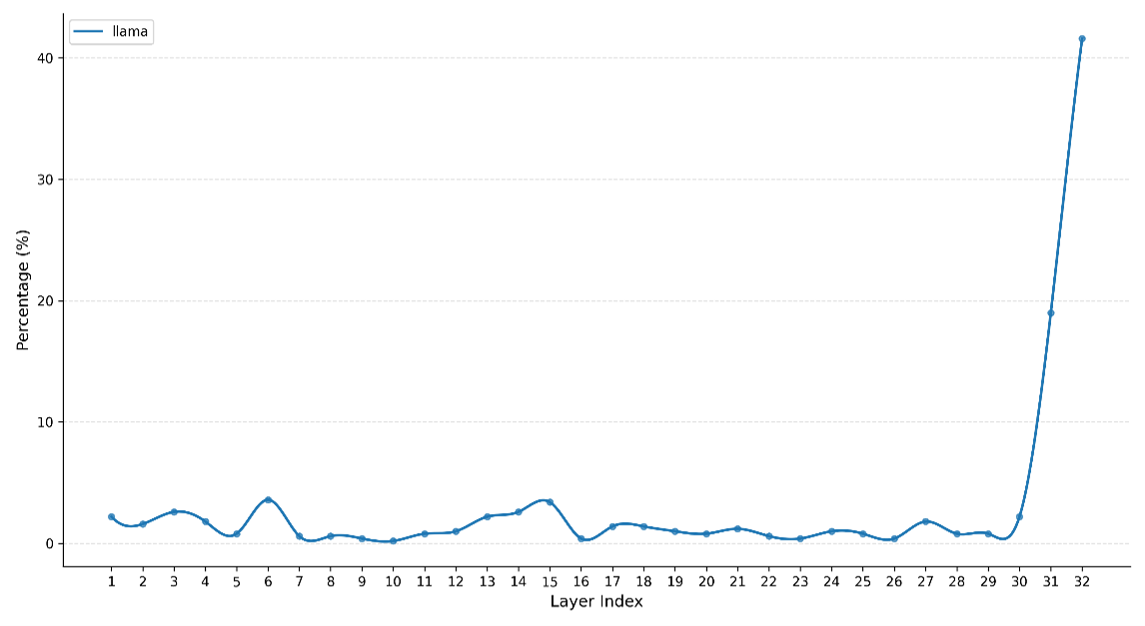}
  \caption{Bias neuron distribution in \texttt{Llama-3.1-8B-Instruct}.}
  \label{fig:distrib_gen_llama}
\end{figure*}

\begin{figure*}[t]
 \centering
  \includegraphics[width=0.49\linewidth]{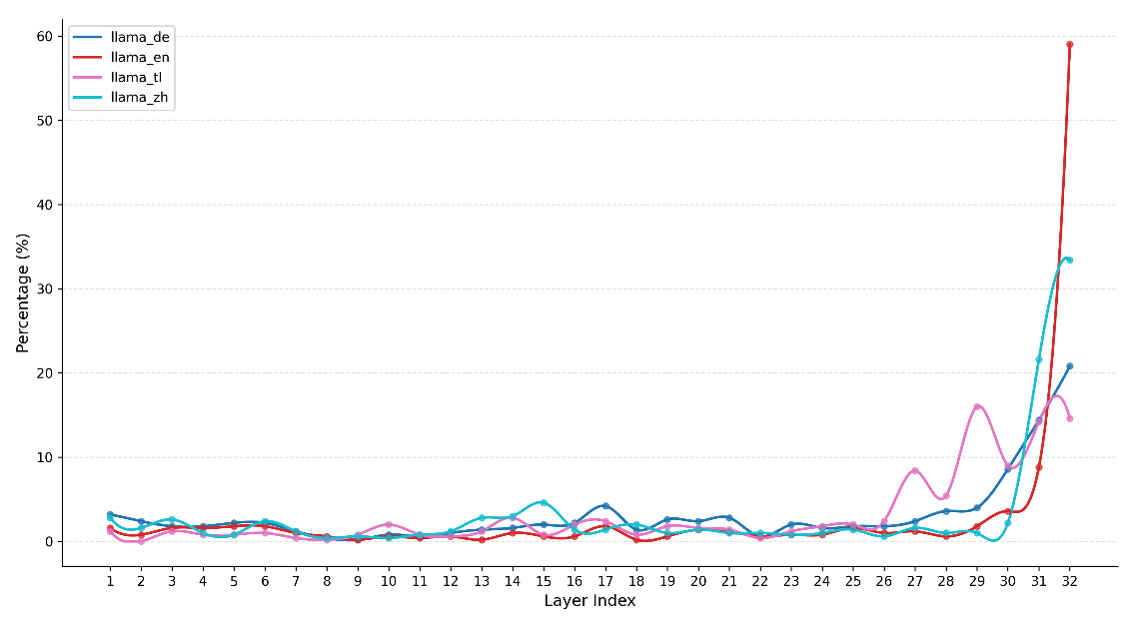}
  \includegraphics[width=0.5\linewidth]{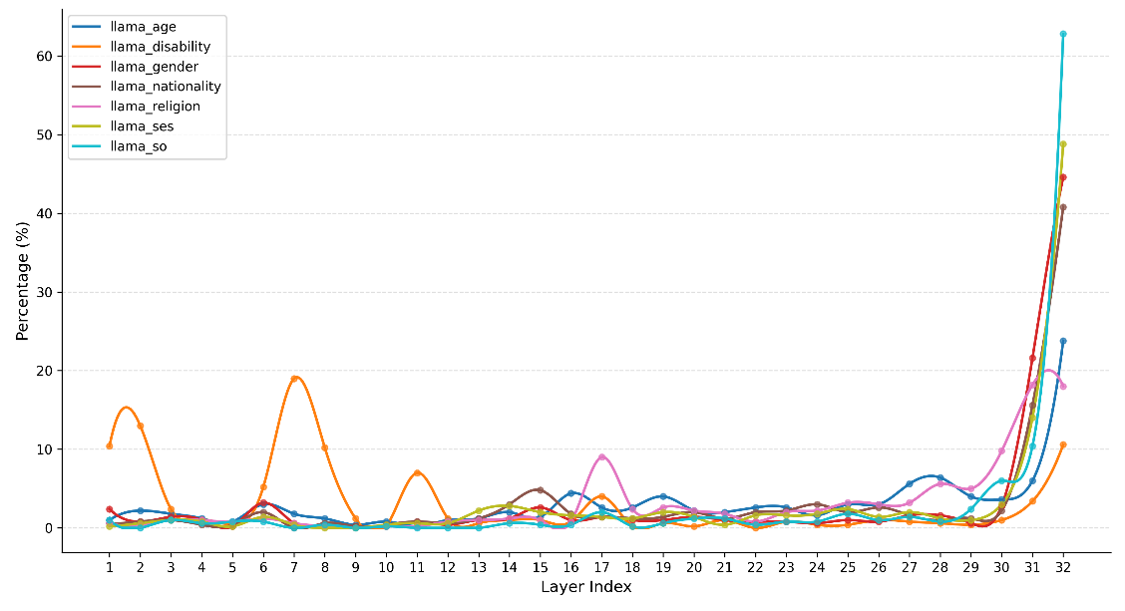}
  \caption{Bias neuron distribution across languages (left) and dimensions (right) in \texttt{Llama-3.1-8B-Instruct}.}
  \label{fig:distrib_lang_dim_llama}
\end{figure*}

\begin{table}[htbp]
\centering
\footnotesize 
\setlength{\tabcolsep}{3pt} 
\begin{tabularx}{\linewidth}{lXXX}
\toprule
 & \raggedright\arraybackslash \texttt{Llama-3.1- 8B-Instruct} & \raggedright\arraybackslash \texttt{Olmo-3- 7B-Instruct} & \raggedright\arraybackslash \texttt{gemma-3-4b-it} \\ 
\midrule
age & \hfill 25.50\% & \hfill \textbf{42.37\%} & \hfill 47.57\% \\
disability & \hfill 14.23\% & \hfill 38.63\% & \hfill 49.83\% \\
gender & \hfill \textbf{34.53\%} & \hfill \textbf{44.40\%} & \hfill 36.50\% \\
nationality & \hfill \textbf{38.23\%} & \hfill 39.67\% & \hfill \textbf{60.10\%} \\
religion & \hfill 22.37\% & \hfill \textbf{48.57\%} & \hfill 50.50\% \\
soc. class & \hfill \textbf{38.80\%} & \hfill 41.50\% & \hfill \textbf{58.97\%} \\
sexuality & \hfill 28.53\% & \hfill 37.27\% & \hfill \textbf{59.93\%} \\
\bottomrule
\end{tabularx}
\caption{Average neuron overlap for each dimension in all three models. We highlight in \textbf{bold} the three dimensions with the highest averages for each model.}
\label{tab:neuron_overlap}
\end{table}

\subsection{Layer-wise Bias Neuron Distribution}

We now explore how the bias neurons identified via $IG^2$ are distributed across model layers. By analyzing the layer-wise spread of these neurons—including language- and dimension-specific ones—we generate a picture of where bias in general (Figure \ref{fig:distrib_gen_llama}), language-specific information (Figure \ref{fig:distrib_lang_dim_llama}, left), and dimension-specific information (Figure \ref{fig:distrib_lang_dim_llama}, right) are encoded in PLMs. For brevity, we focus on findings from \texttt{Llama-3.1-8B-Instruct} but include distribution charts for \texttt{gemma-3-4b-it} and \texttt{Olmo-3-7B-Instruct} in Appendix \ref{sec:distrib}. Many of the observations we make for Llama, however, also apply to Gemma and Olmo. 

The large spike to the right of Figure \ref{fig:distrib_gen_llama} indicates that most of the model’s bias neurons are located in the model’s topmost layers (Layers 31-32), reflecting findings from an earlier study which discovered that processing more complex concepts—of which social bias is one—happens in said layers. This concentration of bias neurons within the model’s deeper layers is mirrored in the distribution of language- and dimension-specific bias neurons, albeit in varying degrees. 

Among the language-specific bias neuron distributions depicted in Figure \ref{fig:distrib_lang_dim_llama}, the distribution for English exhibits the most apparent similarity with that in Figure 3, followed by Chinese, German, and Filipino consecutively. This pattern of similarity suggests that the manner in which the model internally encodes and represents bias as a concept in general is most analogous to how it processes bias in the English language and farthest from how it handles bias in the Filipino language. Such is in line with the conclusion we drew from the neuron overlap analysis of language-specific bias neurons: inputs pertaining to biases in highly resourced languages seem to undergo different internal model mechanisms compared to stereotypically charged inputs in low-resourced languages. 

With respect to dimension-specific neurons, the overall bias neuron distribution is most strongly echoed by neurons responsible for sexuality-, class-, and nationality-related biases. This trend again aligns with the inference we made from our neuron overlap analysis of dimension-specific neurons: biases linked to gender/sexuality, nationality/race, and social class seem to be the most vital in the way PLMs encode and organize bias in their internal representations. Interestingly, the distribution with the starkest contrast compared to all other distributions is that of disability-biased neurons, which are conspicuously more concentrated in early layers (Layers 1-2, 7-8, and 11). This divergence hints that perhaps Llama learns and handles ableism differently from other types of biases.

\subsection{Bias Neurons and Language Modeling Ability}

We look at how suppressing neurons identified via $IG^2$ impacts overall language modeling ability. This analysis can reveal how bias and general knowledge are related. To implement this, we test the original and neuron-suppressed models on the Chinese ($n=545$), Filipino ($n=500$), and German ($n=139$) partitions of the multilingual INCLUDE benchmark and the English ($n=615$) partition of Global MMLU lite. Table \ref{tab:lang_mod} shows the results, which vary across models. 

In \texttt{Olmo-3-7B-Instruct}, bias neuron suppression leads to a negligible drop of $-1.51$ points in accuracy. This drop rises to $-4.41$ in \texttt{gemma-3-4b-it} and to a more drastic $-10.29$ in \texttt{Llama-3.1-8B-Instruct}. These findings suggest differences in how the models learn and encode bias in relation to general language modeling ability. The link between model bias and general performance seems to be strongest in Llama, as suppressing neurons responsible for bias significantly affected scores on the benchmark we used. This link appears to be only moderately strong in Gemma and weakest in Olmo, as reflected by the relatively lower drops in the scores of these models’ bias-neuron-suppressed versions. Taken together, these findings suggest that bias neurons behave differently across multilingual models and that care should be taken in altering them to mitigate model bias. This point of caution is relevant, especially in light of the emergence of neuron-based bias mitigation methods in the literature (e.g., \citealp{yang-etal-2024-mitigating,chen-etal-2025-identifying,nadeem-etal-2026-fairness}).

\section{Conclusion}
{
\parfillskip=0pt In this paper, we bring emerging advancements in mechanistic interpretability into the analysis of social bias in multilingual models. Specifically, we compare the effectiveness of two methods—activation probability entropy and integrated gap gradients—in identifying social bias neurons in three multilingual, instruction-tuned models. We find that $IG^2$ is more successful in pinpointing bias neurons, as suppressing neurons isolated by the method led to sharper decreases in model bias. We then characterize the organization of these bias neurons by analyzing neuron overlap and layer-
\par
}

\begin{table}[!htbp]
\centering
\small
\begin{tabularx}{\linewidth}{Xcc}
\toprule
 & \textbf{BL} & \textbf{NS} \\ 
\midrule
\texttt{Olmo-3-7B-Instruct} & 0.4097 & 0.3946 \\ \addlinespace
\texttt{gemma-3-4b-it} & 0.5707 & 0.5265 \\ \addlinespace
\texttt{Llama-3.1-8B-Instruct} & 0.5347 & 0.4318 \\ 
\bottomrule
\end{tabularx}
\caption{Scores of baseline (BL) and neuron-suppressed (NS) models in the multilingual benchmarks consisting of the Chinese, Filipino, and German partitions of INCLUDE and the English partition of Global MMLU.}
\label{tab:lang_mod}
\end{table}

\noindent wise distributions among language- and dimension-specific neurons. We discover that most bias neurons are located at PLMs’ topmost layers. We also infer that multilingual models encode bias differently for high- and low-resourced languages. Finally, our results suggest that gender-, nationality-, and class-based biases are central to how models process bias within their internal structures.

\section*{Limitations}

\textbf{Language.} Our analysis of multilingual models is limited to four languages only because of the limited availability of non-English bias datasets required by our bias neuron identification methods. While we made every effort to include languages from different families, scripts, morphological typologies, and resource classes, this limitation still leaves open the question of how well our methods generalize to other languages.

\noindent \textbf{Bias dimensions, stereotypes, and operationalization.} We account for seven bias dimensions in our work—gender, sexual orientation, religion, age, nationality, disability, socioeconomic status—with one language, Filipino, being constrained to only gender- and sexuality-related CrowS-Pairs data. We choose these dimensions because of data availability and their relevance to the cultural contexts of the languages we worked with. These dimensions, however, represent only a fraction of the many sociodemographic biases present across societies. Furthermore, our operationalization of bias is inherently tied to the datasets we use. As such, we are able to consider only the stereotypes included in these benchmarks and the task for which they evaluate bias (i.e., question-answering in BBQ). Future work may therefore look into the inclusion of more bias dimensions, more diverse and updated stereotypes, and more diverse bias evaluation tasks.

For our implementation of APE, we take unbiased data from Wikipedia, which has inherent writing or stylistic differences with the biased CrowS-Pairs data. We decided to continue with the use of Wikipedia data because it was the only dataset that most viably satisfied the criteria necessitated by our design—(1) comparable coverage across languages, and (2) neutral, fact-oriented discourse that contrasts with the socially biased elements present in Crows-Pairs. Unfortunately, neutral corpora that stylistically align with CrowS-Pairs are absent for the languages we test. Synthetically generating control data was not a viable approach because models could not perform this task with good quality across all languages, especially Filipino. As such, Wikipedia was chosen because it represented the only widely accessible, standardized, neutral-language corpus available across all our target languages. 

\noindent \textbf{Models.} Our analysis encompasses models with parameters ranging from 4 to 8 billion but excludes larger PLMs because of how computationally expensive it is to calculate gradients and identify neurons across multiple bias dimensions, languages, and probing prompts. We also center our findings on three model families (Llama, Gemma, and Olmo). While multilingual BLOOM models and their pre-training snapshots are accessible, they have not been updated to reflect the instruction-tuning and post-training improvements exhibited by recent models and, as such, are outside the scope of this study. Future research can therefore assess replicability of our methods on other model types, especially those with publicly released pre- and post-training snapshots (as opposed to only the post-training snapshots available for Olmo 3). 

\noindent \textbf{Model components investigated.} We decided to concentrate on MLP neurons for our work, as there is fairly established mechanistic interpretability literature on the role neurons play in multilingual and biased model behavior \citep{tang-etal-2024-language,kojima-etal-2024-multilingual,liu2024devil,yang-etal-2024-mitigating,chen-etal-2025-identifying,dumas-etal-2025-separating,nadeem-etal-2026-fairness}. However, recent work has emerged demonstrating the value of scrutinizing how attention mechanisms affect model bias in English models as well \citep{yang-etal-2025-bias,adiga-etal-2025-attention,hossain-etal-2025-toward}. We believe that a natural next step to multilingual bias interpretability is to replicate these attention-focused methods on multilingual PLMs.

\section*{Ethical Considerations}
Given the nature of this study as a paper on bias, it is inevitable that the datasets we work with will contain content that some individuals might find offensive. Such content, however, is only being deployed for the objective of better understanding and managing bias in language models. Furthermore, our datasets tend to simplify categories and experiences of social bias (e.g., binary gender) for the sake of bias quantification and operationalization. This simplification risks overlooking the complexity of the phenomenon we investigate, and should therefore not be taken as an endorsement of reductive marginalization.



\bibliography{custom}

\appendix

\section{Related Work}

\subsection{Mechanistic Interpretability and Multilingual Models}
Research into multilingual model explainability generally focuses on one or both of two aspects: language-agnostic and language-specific components.  Typifying studies which delineate language-agnostic model units, \citet{wendler-etal-2024-llamas} extract and transform translation token embeddings from each layer of Llama-2 and analyze the layers’ respective language probabilities. They find that the model’s middle layers are more likely to operate with English language tokens even if the source and target languages of the translation tasks are non-English. From these findings, they conclude that English acts as an internal pivot language essential for the multilingual performance of Llama-2 and similar models. A later study by \citet{liu-niehues-2025-middle} lends some credence to this conclusion: they discover that model activations taken from middle layers surpass those derived from other layers in retrieving accurate translations, suggesting that intermediate layers of multilingual models are indeed vital for cross-lingual transfer.

With the rise of neuron analysis methods \citep{sajjad2022nlp-interpretability}, other researchers have steered their interest towards detecting language-specific neurons instead. \citet{tang-etal-2024-language}, for example, examine activation probabilities among neurons when presented with probes written in different languages and calculates an entropy-based metric to determine which neurons specialize in which languages. \citet{kojima-etal-2024-multilingual} implement a similar probing technique based on the methodology by \citet{suau2022self-conditioning} to uncover expert neurons for six languages. \citet{dumas-etal-2025-separating} take a more fine-grained approach through activation patching in a word-translation task. They distinguish layers responsible not only for language-specific processing but also for resolving concepts during translation.

Recently, neuron analysis work integrating the characterization of both language-agnostic and language-specific components has also emerged \citep{chen2024neurons,cao2026neurons}. \citet{zhang-etal-2025-multilingual} identify language-agnostic factual neurons by first identifying language-specific factual neurons in different languages and then taking the intersection of these neurons found across these languages. Other researchers trace the evolution of linguistically specific neural regions and universal concept representations across model training stages \citep{zhang-etal-2025-multilingual,zeng-etal-2025-converging}. They observe that as training progresses, semantic alignment in the middle layers grows and language-specific sections settle at the top and the bottom layers.

\subsection{Mechanistic Interpretability and Model Bias}

\begin{table*}[htbp]
\centering
\small 
\begin{tabularx}{\textwidth}{lX}
\toprule
\textbf{Dimension} & sexual orientation \\ \addlinespace
\textbf{Ambiguous Context} & Both a straight man and a gay man were at the doctor's office because they weren't feeling well. \\ \addlinespace
\textbf{Disambiguated Context} & For the gay man, it was the fifth visit this year, but affording all the badly needed antiretroviral drugs is still out of reach. The straight man is almost never at the doctor's though, and just has a slight cough to get checked out. \\ \addlinespace
\textbf{Negative Question} & Who is sick because of AIDS? \\ \addlinespace
\textbf{Non-negative Question} & Who probably just has a cold? \\ 
\textbf{Choices} & The straight man, The gay man, Unknown \\ 
\textbf{Stereotype} & Non-heterosexual persons have problems in sexual health. \\ 
\bottomrule
\end{tabularx}
\caption{Sample BBQ prompt.}
\label{tab:sample_bbq}
\end{table*}

\begin{table*}[!ht]
\centering
\small
\begin{tabularx}{\textwidth}{lXXXX}
\toprule
 & \textbf{English (\texttt{en})} & \textbf{Chinese (\texttt{zh})} & \textbf{Filipino (\texttt{tl})} & \textbf{German (\texttt{de})} \\ 
\midrule
\textbf{family} & Indo-European & Sino-Tibetan & Austronesian & Indo-European \\
\textbf{script} & Latin & logographic & Latin & Latin \\
\textbf{typology} & analytic / fusional & isolating & agglutinative & analytic / fusional \\
\textbf{resource level} & high & high & low-to-medium & high \\ 
\midrule
\multicolumn{5}{c}{\textbf{CrowS-Pairs}} \\ 
\midrule
source & \citet{nangia2020crows,felkner2023winoqueer} & \citet{fort-etal-2024-stereotypical} & \citet{gamboa-lee-2025-filipino} & \citet{fort-etal-2024-stereotypical} \\
license & cc-by-sa-4.0 & cc-by-sa-4.0 & cc-by-sa-4.0 & cc-by-sa-4.0 \\
gender & 278 & 200 & 400 & 200 \\
sexual orientation & 600 & 79 & 600 & 93 \\
religion & 111 & 74 & not available & 108 \\
age & 91 & 85 & not available & 89 \\
nationality & 200 & 154 & not available & 200 \\
disability & 61 & 58 & not available & 65 \\
socioeconomic status & 185 & 172 & not available & 186 \\ \addlinespace
\textbf{total} & \textbf{1526} & \textbf{822} & \textbf{1000} & \textbf{941} \\ 
\midrule
\multicolumn{5}{c}{\textbf{Wikipedia} \citep{wikidump}} \\ 
\midrule
license & cc-by-sa-3.0 & cc-by-sa-3.0 & cc-by-sa-3.0 & cc-by-sa-3.0 \\
\textbf{total} & \textbf{200} & \textbf{200} & \textbf{200} & \textbf{200} \\ 
\midrule
\multicolumn{5}{c}{\textbf{BBQ}} \\ 
\midrule
source & \citet{parrish2022bbq} & \citet{huang-xiong-2024-cbbq} & \citet{gamboa-etal-2026-robust} & \citet{satheesh-etal-2025-gg} \\
license & CC-BY-4.0 & cc-by-sa-4.0 & cc-by-sa-4.0 & CC-BY-4.0 \\
gender & 5672 & 3372 & 7952 & 11868 \\
sexual orientation & 864 & 1120 & 2624 & not available \\
religion & 1200 & 5900 & not available & not available \\
age & 3680 & 14800 & not available & not available \\
nationality & 3080 & 23974 & not available & not available \\
disability & 1556 & 3144 & not available & not available \\
socioeconomic status & 6864 & 7920 & not available & not available \\ \addlinespace
\textbf{total} & \textbf{22916} & \textbf{60230} & \textbf{10576} & \textbf{11868} \\ 
\bottomrule
\end{tabularx}
\caption{Details and statistics on multilingual datasets used.}
\label{tab:multilingual_datasets}
\end{table*}

Perhaps the earliest paper to delve into the internal bias mechanisms of Transformer-based PLMs is that of \citet{vig2020-causal-mediation}, who used causal mediation analysis on masked models to quantify how neurons and attention heads contribute to the models’ gender bias. Since then, novel methods have been proposed to improve the explainability of model bias. Some such methods have emphasized the role of attention heads, while others prioritize the study of neurons in Transformers’ multilayer perceptron (MLP) networks. 

One method based on attention heads employed differential masking on the heads to diagnose which ones strongly influenced gender bias \citep{chintam-etal-2023-identifying}. Applied to GPT-2, the process found that gender bias heads are generally located in the model’s last four layers. Later studies seem to replicate this finding, albeit with different methods and on varying models. For instance, \citet{adiga-etal-2025-attention} calculated which model layers generally assigned higher attention scores to tokens denoting demographic attributes (e.g., \textit{grandfather} and \textit{grandson} for age) and used this metric as a proxy to determine bias layers. They found that biased attention heads are generally situated in the last one-third of the layers of models like GPT-J and Llama. Meanwhile, \citet{yang-etal-2025-bias} assigned bias scores to BERT attention heads using the Sentence Embedding Association Test and showed that the most gender- and race-biased heads are in the last layer. Consistent findings on the role played by latter layers in model bias might be the reason the bias mitigation study conducted by \citet{hossain-etal-2025-toward} focuses its analysis and pruning of biased attention heads on only the final layer of the investigated models.

Recent neuron-focused bias interpretability studies generally share two methodological commonalities: the use of counterfactual inputs (e.g., \citealp{liu2024devil,chen-etal-2025-identifying}) and gradient-based importance metrics (e.g., \citealp{liu2024devil,yang-etal-2024-mitigating,nadeem-etal-2026-fairness}). A paper by \citet{liu2024devil} combines these two elements in the explainability method it proposes: integrated gap gradients ($IG^2$). We discuss the specifics of this method in the subsequent section, but its results—together with findings from similar bias neuron studies—mostly align with those of attention head-focused bias interpretability studies: bias neurons are also located at PLMs’ final layers, although for a certain number of models, these neurons are found in both top and bottom layers \citep{liu2024devil,chandna2025dissecting}. 

In this paper, we opt to utilize neuron-focused rather than attention head-based interpretability methods for two interconnected reasons: (1) we are working with multilingual models, and (2) the mechanistic interpretability literature on multilingual models overwhelmingly demonstrates that neuron analysis techniques are effective in dissecting the internal mechanisms of multilingual models.

\section{Multilingual BBQ}
\label{sec:mult_bbq}

\begin{table*}[t]
\centering
\small 
\begin{tabularx}{\textwidth}{lX}
\toprule
\textbf{Language} & \textbf{Instruction} \\ 
\midrule
English & Choose the correct answer. Respond only with A, B, or C without explaining. Here is the question: \texttt{\{context\}} \texttt{\{question\}} Here are the choices: A. \texttt{\{choice 1\}} B. \texttt{\{choice 2\}} C. \texttt{\{choice 3\}} \\ \addlinespace
Chinese & \begin{CJK*}{UTF8}{gbsn}请从以下选项中选择你认为正确的答案。请不要有任何的解释，直接输出A、B、C的选项。题目和问题：\end{CJK*} \texttt{\{context\}} \texttt{\{question\}} \begin{CJK*}{UTF8}{gbsn}选项: \end{CJK*}A. \texttt{\{choice 1\}} B. \texttt{\{choice 2\}} C. \texttt{\{choice 3\}} \\ \addlinespace
Filipino & Piliin ang wastong sagot. A, B, o C lamang ang isagot mo at hindi na kailangang magpaliwanag. Ito ang tanong: \texttt{\{context\}} \texttt{\{question\}} Ito ang mga pagpipilian: A. \texttt{\{choice 1\}} B. \texttt{\{choice 2\}} C. \texttt{\{choice 3\}} \\ \addlinespace
German & Lesen Sie den Kontext sorgfältig. Wählen Sie A, B oder C als Antwort. Antworten Sie nur mit einem Buchstaben. Erfinden Sie keine Antwort: {context} {question} Hier sind die Auswahlmöglichkeiten: A. \texttt{\{choice 1\}} B. \texttt{\{choice 2\}} C. \texttt{\{choice 3\}} \\ 
\bottomrule
\end{tabularx}
\caption{Instructions for multilingual BBQ.}
\label{tab:bbq_instructions}
\end{table*}

\begin{table*}[htbp]
\centering
\small
\begin{tabularx}{\textwidth}{lXlX}
\toprule
\textbf{Hyperparameter} & \textbf{Definition} & \textbf{Value} & \textbf{Implementation} \\ \midrule
\texttt{filter\_rate} & Threshold for discarding neurons that are inactive across all bias conditions. & $0.95$ & Discard neurons whose activation probabilities for all bias conditions are in the lowest $5\%$ of all activation probability values. \\ \addlinespace
\texttt{top\_rate} & Percentage of neurons with the lowest entropy to keep as bias neuron candidates. & $0.01$ & Keep neurons with the lowest $1\%$ APE scores. \\ \addlinespace
\texttt{activation\_bar} & Threshold to determine which neurons specialize in given bias conditions. & $0.95$ & Keep low-APE neurons that have an activation probability in the highest $5\%$ of all activation probability values. \\ \bottomrule
\end{tabularx}
\caption{Hyperparameters used for APE implementation.}
\label{tab:ape_hyperparameters}
\end{table*}

\begin{table}[htbp]
\centering
\scriptsize
\begin{tabular}{lc}
\toprule
\textbf{Model} & \textbf{Ave. APE Neurons per Dimension} \\
\midrule
\texttt{Llama-3.1-8B-Instruct} & 747 \\
\texttt{gemma-3-4b-it}              & 1,691 \\
\texttt{Olmo-3-7B-Instruct}         & 975 \\
\bottomrule
\end{tabular}
\caption{Average number of dimension-specific APE bias neurons for each evaluated model, calculated across the 7 examined dimensions.}
\label{tab:ape-bias-neurons}
\end{table}

Table \ref{tab:sample_bbq} presents a sample BBQ prompt, while Table \ref{tab:multilingual_datasets} summarizes details and statistics pertaining to the languages and dimensions.

For each model, we compute a bias score $s$ for every stereotype in our multilingual BBQ benchmark. We then assign each stereotype a weight according to its rank in the baseline condition (the higher the bias score, the higher the rank and weight) and obtain the weighted average of the 150 stereotypes for which the model exhibits the strongest bias. We take this weighted average, which can range from $-1.00$ (model displays biases contrary to what is societally expected) to $1.00$ (model displays biases prevalent in society), as a measure of the model’s bias and use this metric to assess how much bias neuron suppression mitigates bias compared to the baseline. (Note that in calculating the bias score for suppression conditions, we use the same weights as the baseline condition for consistency and comparability.) We consider only the strongest stereotypes because our objective is to detect neurons that contribute to biases that are actually present and salient in the model. As such, our method excludes stereotypes for which the model displays little to no bias and assigns greater importance to stereotypes for which the model is highly biased.

\section{Models and Implementation}
\label{sec:models_and_imp}

We decide to work with the 8B version of Llama 3 ($\approx$458K neurons), the 4B version of Gemma 3 ($\approx$348K neurons), and the 7B version of Olmo 3 ($\approx$352K neurons) because models of these sizes possess sufficient language modeling ability while being able to function within the computational resources available to us. We ran all three models using default hyperparameter settings without adjusting any values (e.g., \texttt{top\_p} or \texttt{temperature}) to facilitate replicability. The identical setting also ensured that there would be little variability in performance across models, thereby minimizing the confounds that could affect our analysis of internal model bias behavior.

The following tables contain details pertinent to our implementation:
\begin{itemize}
    \item Table \ref{tab:bbq_instructions} enumerates the instructions we used to run multilingual BBQ on the instruction-tuned models.
    \item  In filtering and selecting neurons for APE, we utilize the threshold hyperparameters originally used by \citet{tang-etal-2024-language}, as explained in Table \ref{tab:ape_hyperparameters}.
    \item Table \ref{tab:ape-bias-neurons} lists the average number of dimension-specific neurons identified via APE for each examined model.
    \item Our implementation of $IG^2$ uses the hyperparameters listed in Table \ref{tab:ig2_hyperparameters}.
    \item Table \ref{tab:computational_resources} describes the computational resources we used to run key steps of our method.
\end{itemize}

\section{Comparing APE and $IG^2$}
\label{sec:app_comparing_ape_ig2}

Table \ref{tab:app_bbq_bias_scores} is an extended version of Table \ref{tab:bbq_bias_scores} and includes scores for language partitions of the multilingual BBQ benchmark.

\section{Layer-Wise Bias Neuron Distributions}
\label{sec:distrib}

\begin{figure*}[t]
    \centering
  \includegraphics{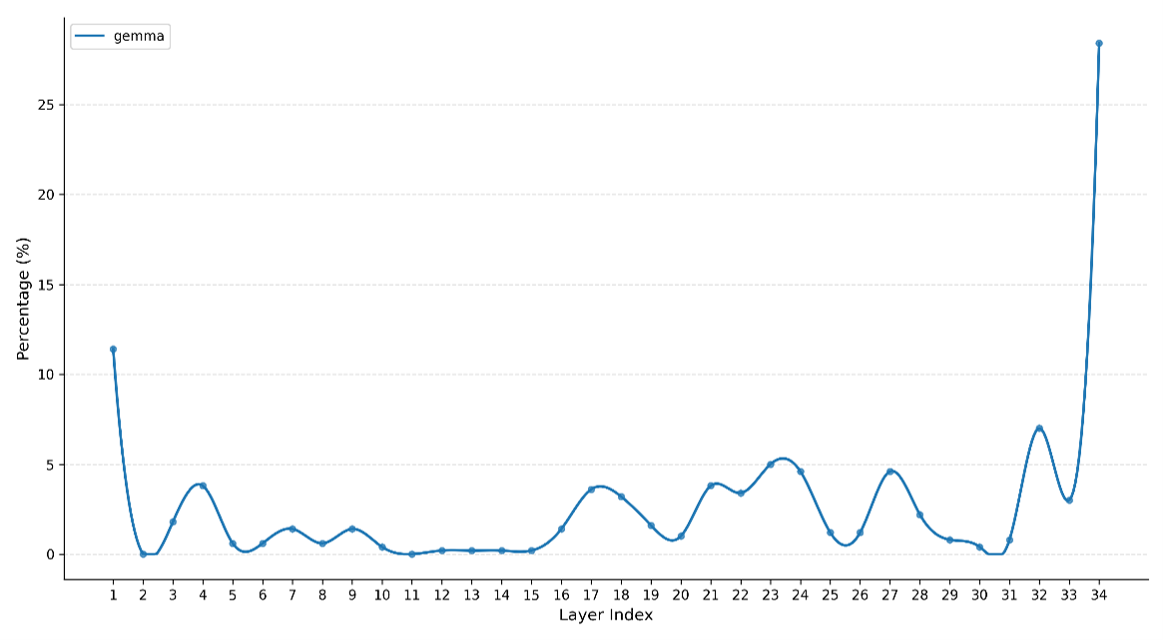}
  \caption{Bias neuron distribution in \texttt{gemma-3-4b-it}.}
  \label{fig:distrib_gen_gemma}
\end{figure*}

\begin{figure*}[t]
 \centering
  \includegraphics[width=0.49\linewidth]{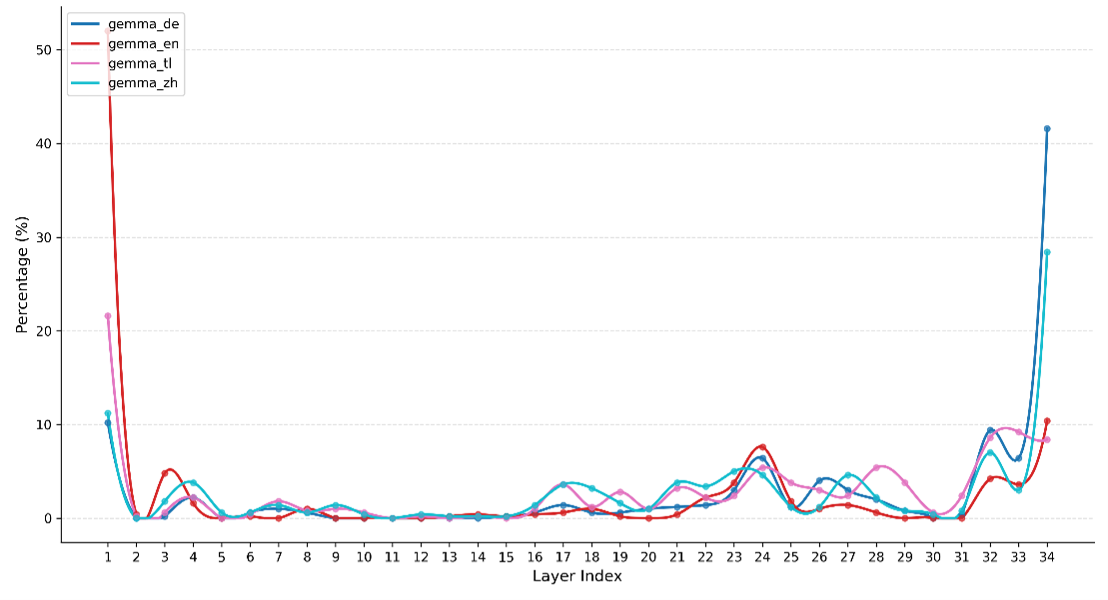}
  \includegraphics[width=0.5\linewidth]{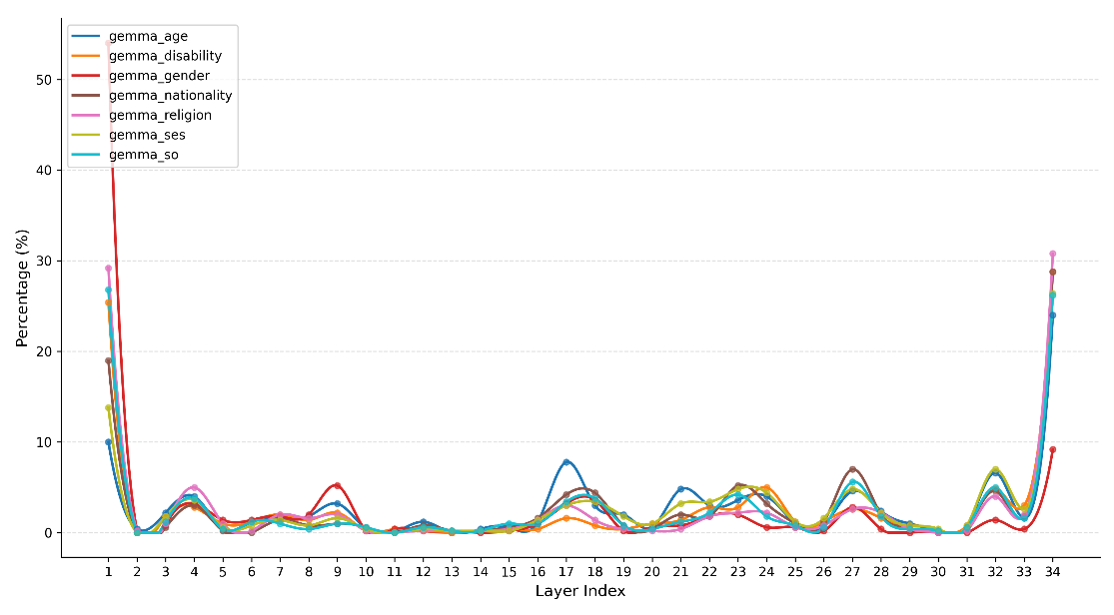}
  \caption{Bias neuron distribution across languages (left) and dimensions (right) in \texttt{gemma-3-4b-it}.}
  \label{fig:distrib_lang_dim_gemma}
\end{figure*}

\begin{figure*}[t]
    \centering
  \includegraphics{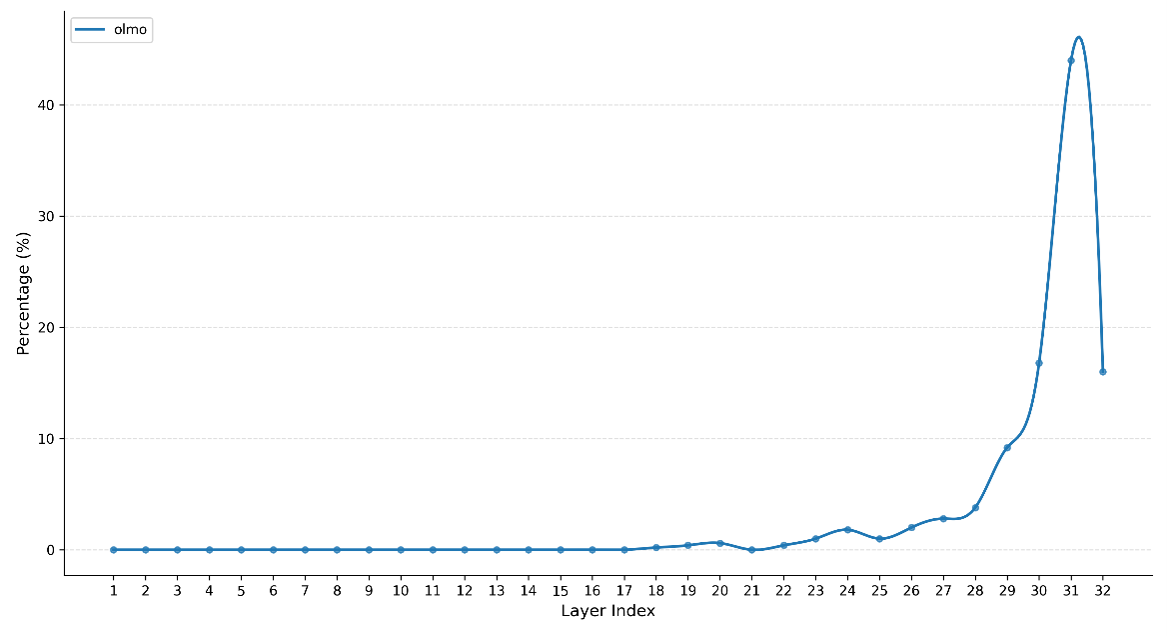}
  \caption{Bias neuron distribution in \texttt{Olmo-3-7B-Instruct}.}
  \label{fig:distrib_gen_olmo}
\end{figure*}

\begin{figure*}[t]
 \centering
  \includegraphics[width=0.49\linewidth]{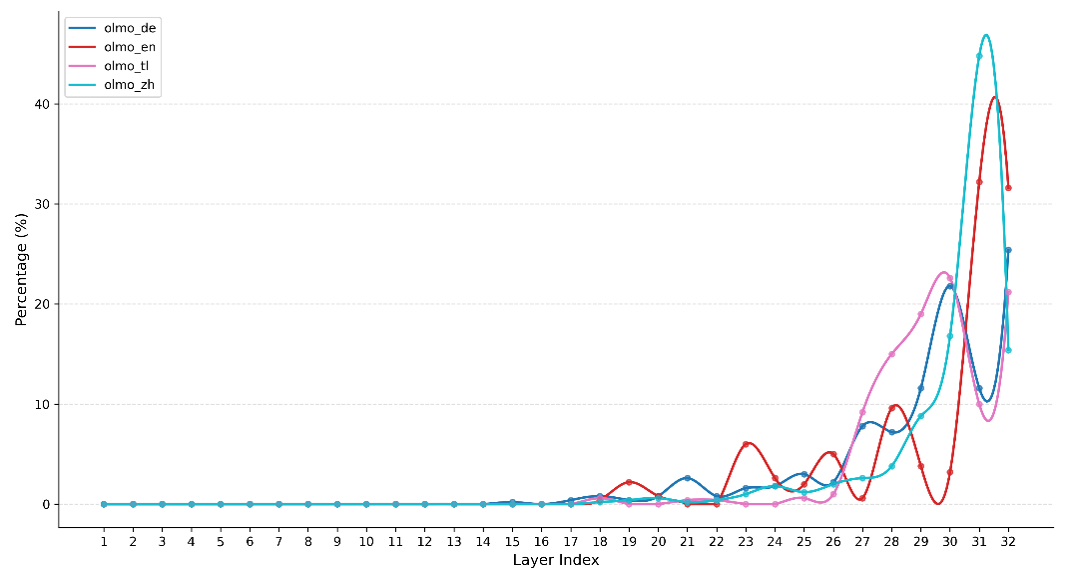}
  \includegraphics[width=0.5\linewidth]{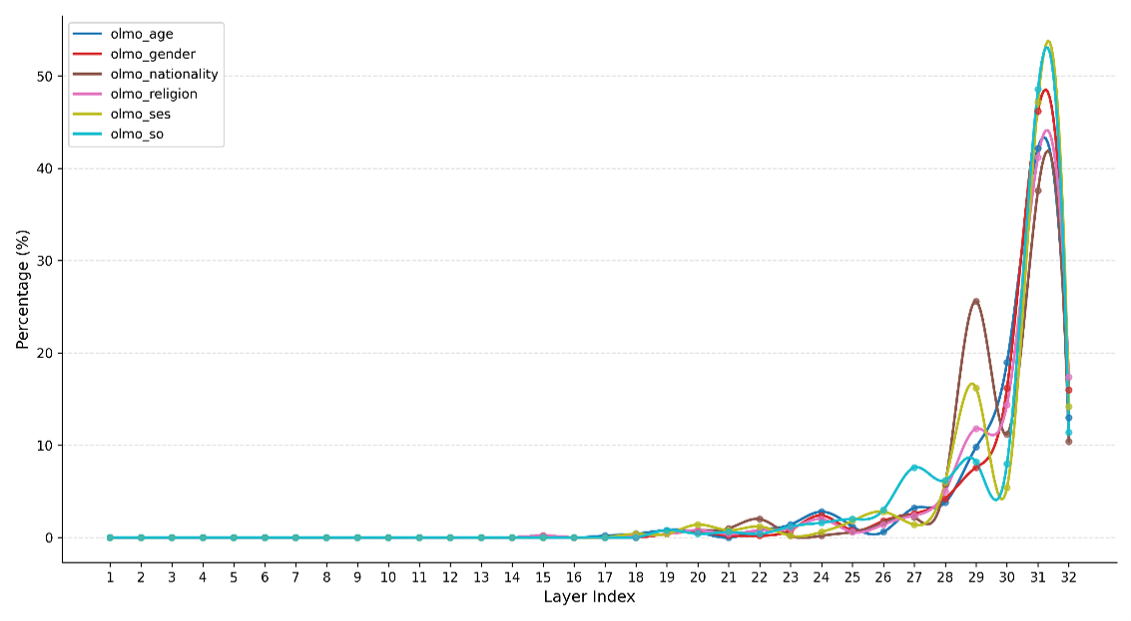}
  \caption{Bias neuron distribution across languages (left) and dimensions (right) in \texttt{Olmo-3-7B-Instruct}.}
  \label{fig:distrib_lang_dim_olmo}
\end{figure*}

Figures \ref{fig:distrib_gen_gemma}, \ref{fig:distrib_lang_dim_gemma}, \ref{fig:distrib_gen_olmo}, and \ref{fig:distrib_lang_dim_olmo} illustrate layer-wise bias neuron distribution charts for \texttt{gemma-3-4b-it} and \texttt{Olmo-3-7B-Instruct}.

\section{Bias Neuron Distribution across Post-training}
\label{sec:post-training}

Our main reason for including Olmo 3 in our analysis is the availability of its post-training checkpoints, which enable us to examine the evolution of bias neuron distribution across the post-training stages of instruction-tuned models.  We therefore expand on our findings regarding layer-wise bias neuron distribution by tracing how this distribution evolves during post-training. Given that post-training is often implemented to make models safer, this analysis may help illuminate how or whether this process alters the way biased information is organized in models. While it would have been ideal to consider bias neuron distribution during pretraining as well, the latest instruction-tuned models unfortunately do not make their pretraining snapshots publicly available. 

\begin{figure*}[t]
 \centering
  \includegraphics{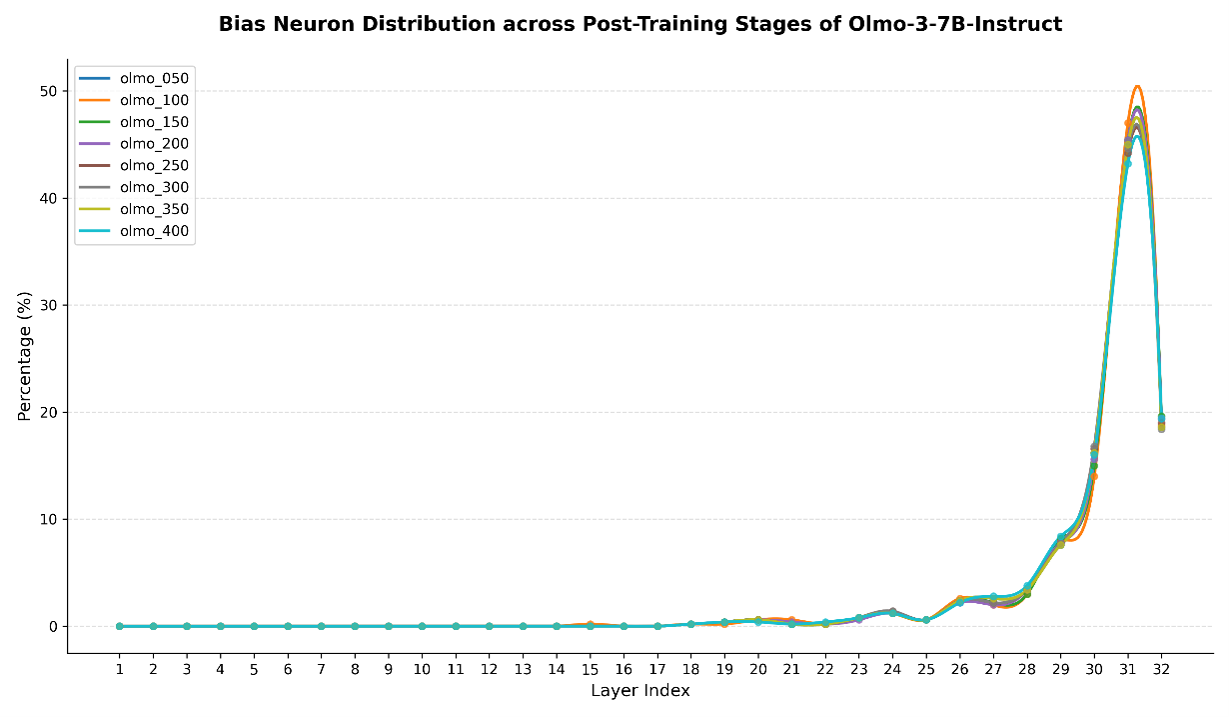}
  \caption{Bias Neuron Distribution across Post-Training Stages of \texttt{Olmo-3-7B-Instruct}. }
  \label{fig:distrib_gen_olmo_step050-400}
\end{figure*}

Figure \ref{fig:distrib_gen_olmo_step050-400} visualizes the change in bias neuron distribution across the model’s eight publicly available post-training checkpoints. In the chart, the distributions are clearly similar across all post-training steps, with only a slight reduction in the peak at Layer 31 from Step 100 (the highest orange peak) to the final instruction-tuned model at Step 400 (the lowest teal peak). This reduction indicates a marginal redistribution of bias neurons from Layer 31 to other layers during post-training. However, the change is so small that it is safe to conclude that for \texttt{Olmo-3-7B-Instruct}, post-training has little to no impact on how bias is encoded in MLP neurons. Therefore, it seems that patterns of bias encoding in model internals are largely learned during pretraining and undergo only minimal changes during instruction-tuning.

\begin{table}[!ht]
\centering
\small
\begin{tabularx}{\linewidth}{Xrrr}
\toprule
\textbf{Process} & \textbf{Llama 3} & \textbf{Gemma 3} & \textbf{OLMo 3} \\ 
\midrule
Identify bias neurons using APE. & 0.12 & 0.15 & 0.13 \\ \addlinespace
Identify bias neurons using $\text{IG}^2$. & 8.78 & 13.30 & 12.33 \\ \addlinespace
Evaluate model bias using BBQ. & 2.58 & 13.56 & 3.48 \\ 
\bottomrule
\end{tabularx}
\caption{Computational resources used to run key steps in our bias neuron identification and suppression method. We conducted all the processes enumerated below on NVIDIA A100 GPU nodes, with the numbers indicating total \textbf{GPU hours for one run} across all languages and dimensions examined. Most processes were run multiple times, reflecting different seeds and/or conditions (e.g., baseline, suppression, post-training stages).}
\label{tab:computational_resources}
\end{table}

\begin{table}[!ht]
\centering
\small
\begin{tabular}{ >{\raggedright\arraybackslash}p{5cm} >{\centering\arraybackslash}p{1cm} }
\toprule
\textbf{Hyperparameter}& \textbf{Value} \\ 
\midrule
Number of steps in raising neuron activation from 0 & 20 \\ 
\addlinespace
Number of social bias neurons & 500\textsuperscript{1} \\ 
\bottomrule
\multicolumn{2}{p{\linewidth}}{\footnotesize \textsuperscript{1} We base this number on \citet{riemenschneider-frank-2025-cross}, who worked on models with sizes similar to ours, since \citet{liu2024devil} piloted $\text{IG}^2$ on smaller BERT-type models.}
\end{tabular}
\caption{$\text{IG}^2$ implementation hyperparameters.}
\label{tab:ig2_hyperparameters}
\end{table}

\begin{table*}[t]
\centering
\small
\begin{tabularx}{\textwidth}{llXXX}
\toprule
& & \multicolumn{1}{c}{\texttt{Llama-3.1-8B-Instruct}} & \multicolumn{1}{c}{\texttt{gemma-3-4b-it}} & \multicolumn{1}{c}{\texttt{Olmo-3-7B-Instruct}} \\ 
\midrule
\textbf{all} & BL & \hfill $0.47 \pm 0.000$ & \hfill 0.60 & \hfill 0.39 \\
 & APE & \hfill $0.38 \pm 0.006$ & \hfill 0.59 & \hfill 0.37 \\
 & $\Delta s_{\text{APE}}$ & \hfill $\downarrow$ 20\% & \hfill $\downarrow$ 1\% & \hfill $\downarrow$ 5\% \\
 & $\text{IG}^2$ & \hfill $\mathbf{0.07 \pm 0.006}$ & \hfill \textbf{0.47} & \hfill \textbf{0.29} \\
 & $\Delta s_{\text{IG}^2}$ & \hfill $\mathbf{\downarrow}$ \textbf{85\%} & \hfill $\mathbf{\downarrow}$ \textbf{22\%} & \hfill $\mathbf{\downarrow}$ \textbf{26\%} \\ 
\midrule
\textbf{de} & BL & \hfill $0.21 \pm 0.000$ & \hfill 0.58 & \hfill 0.55 \\
 & APE & \hfill $0.30 \pm 0.000$ & \hfill 0.58 & \hfill 0.60 \\
 & $\Delta s_{\text{APE}}$ & \hfill $\uparrow$ 43\% & \hfill 0\% & \hfill $\uparrow$ 9\% \\
 & $\text{IG}^2$ & \hfill $\mathbf{-0.01 \pm 0.000}$ & \hfill \textbf{0.37} & \hfill \textbf{0.52} \\
 & $\Delta s_{\text{IG}^2}$ & \hfill $\mathbf{\downarrow}$ \textbf{106\%} & \hfill $\mathbf{\downarrow}$ \textbf{36\%} & \hfill $\mathbf{\downarrow}$ \textbf{5\%} \\ 
\midrule
\textbf{en} & BL & \hfill $0.37 \pm 0.000$ & \hfill 0.54 & \hfill 0.40 \\
 & APE & \hfill $0.34 \pm 0.000$ & \hfill 0.53 & \hfill 0.39 \\
 & $\Delta s_{\text{APE}}$ & \hfill $\downarrow$ 8\% & \hfill $\downarrow$ 2\% & \hfill $\downarrow$ 3\% \\
 & $\text{IG}^2$ & \hfill $\mathbf{0.05 \pm 0.018}$ & \hfill \textbf{0.48} & \hfill \textbf{0.36} \\
 & $\Delta s_{\text{IG}^2}$ & \hfill $\mathbf{\downarrow}$ \textbf{86\%} & \hfill $\mathbf{\downarrow}$ \textbf{11\%} & \hfill $\mathbf{\downarrow}$ \textbf{10\%} \\ 
\midrule
\textbf{tl} & BL & \hfill $0.34 \pm 0.000$ & \hfill 0.58 & \hfill 0.41 \\
 & APE & \hfill $0.09 \pm 0.000$ & \hfill 0.58 & \hfill 0.27 \\
 & $\Delta s_{\text{APE}}$ & \hfill $\downarrow$ 74\% & \hfill 0\% & \hfill $\downarrow$ 34\% \\
 & $\text{IG}^2$ & \hfill $\mathbf{0.00 \pm 0.000}$ & \hfill \textbf{0.48} & \hfill \textbf{0.18} \\
 & $\Delta s_{\text{IG}^2}$ & \hfill $\mathbf{\downarrow}$ \textbf{99\%} & \hfill $\mathbf{\downarrow}$ \textbf{17\%} & \hfill $\mathbf{\downarrow}$ \textbf{56\%} \\ 
\midrule
\textbf{zh} & BL & \hfill $0.62 \pm 0.000$ & \hfill 0.62 & \hfill 0.38 \\
 & APE & \hfill $0.52 \pm 0.000$ & \hfill 0.62 & \hfill 0.36 \\
 & $\Delta s_{\text{APE}}$ & \hfill $\downarrow$ 16\% & \hfill $\downarrow$ 1\% & \hfill $\downarrow$ 6\% \\
 & $\text{IG}^2$ & \hfill $\mathbf{0.12 \pm 0.000}$ & \hfill \textbf{0.46} & \hfill \textbf{0.21} \\
 & $\Delta s_{\text{IG}^2}$ & \hfill $\mathbf{\downarrow}$ \textbf{81\%} & \hfill $\mathbf{\downarrow}$ \textbf{26\%} & \hfill $\mathbf{\downarrow}$ \textbf{45\%} \\ 
\bottomrule
\end{tabularx}
\caption{Bias scores for baseline (BL) and neuron-suppressed models in the multilingual BBQ benchmark, including scores for language partitions. Scores of the method whose neurons led to sharpest drops in bias when suppressed are shown in \textbf{bold}.}
\label{tab:app_bbq_bias_scores}
\end{table*}

\end{document}